\def\year{2022}\relax
\documentclass[letterpaper]{article} 
\usepackage{aaai22}  
\usepackage{times}  
\usepackage{helvet} 
\usepackage{courier}  
\usepackage[hyphens]{url}  
\usepackage{graphicx} 
\usepackage{amsmath}
\usepackage{subcaption,enumitem}
\usepackage{algpseudocode}
\usepackage{xcolor}
\usepackage{multirow}

\newtheorem{definition}{Definition}
\newtheorem{remark}{Remark}
\usepackage[ruled,noend,linesnumbered]{algorithm2e}
\usepackage{booktabs}
\usepackage{multirow}
\usepackage{appendix}
\usepackage{etoolbox}
\usepackage[ruled,noend]{algorithm2e}
\usepackage{booktabs}
\usepackage{multirow}
\usepackage{amssymb}
\BeforeBeginEnvironment{appendix}{\clearpage}

\usepackage{natbib}  
\usepackage{caption} 
\title{
EMVLight: A Decentralized Reinforcement Learning Framework for Efficient Passage of Emergency Vehicles

}
\author{
    Haoran Su, \textsuperscript{\rm 1}\thanks{This work is done through his internship with Siemens Corporation, Technology.}
    Yaofeng Desmond Zhong, \textsuperscript{\rm 2} 
    Biswadip Dey, \textsuperscript{\rm 2}
    Amit Chakraborty \textsuperscript{\rm 2}
    \\
}
\affiliations{
    \textsuperscript{\rm 1}New York University,\\

    \textsuperscript{\rm 2}Siemens Corporation, Technology\\
    haoran.su@nyu.edu, \{yaofeng.zhong, biswadip.dey, amit.chakraborty\}@siemens.com

}

\begin{document}

\maketitle

\begin{abstract}
Emergency vehicles (EMVs) play a crucial role in responding to time-critical events such as medical emergencies and fire outbreaks in an urban area. The less time EMVs spend traveling through the traffic, the more likely it would help save people's lives and reduce property loss. To reduce the travel time of EMVs, prior work has used route optimization based on historical traffic-flow data and traffic signal pre-emption based on the optimal route. However, traffic signal pre-emption dynamically changes the traffic flow which, in turn, modifies the optimal route of an EMV. In addition, traffic signal pre-emption practices usually lead to significant disturbances in traffic flow and subsequently increase the travel time for non-EMVs. In this paper, we propose EMVLight, a decentralized reinforcement learning (RL) framework for simultaneous dynamic routing and traffic signal control. EMVLight extends Dijkstra's algorithm to efficiently update the optimal route for the EMVs in real time as it travels through the traffic network. The decentralized RL agents learn network-level cooperative traffic signal phase strategies that not only reduce EMV travel time but also reduce the average travel time of non-EMVs in the network. This benefit has been demonstrated through comprehensive experiments with synthetic and real-world maps. These experiments show that EMVLight outperforms benchmark transportation engineering techniques and existing RL-based signal control methods.
\end{abstract}

\section{Introduction}
Emergency vehicles (EMVs) are vehicles including ambulances, fire trucks, and police cars, which respond to critical events such as medical emergencies, fire disasters, and criminal activities. Emergency response time is the key indicator of a city's emergency management capability. Reducing response time saves lives and prevents property loss. For instance, the survivor rate from a sudden cardiac arrest without treatment drops 7\% - 10\% for every second elapsed, and there is barely any chance to survive after 8 minutes. EMV travel time, the time interval for an EMV to travel from a rescue station to an incident site, accounts for a major portion of the emergency response time. However, overpopulation and urbanization have been exacerbating road congestion, making it more and more challenging to reduce the average EMV travel time. Records \cite{end-to-end-response-times} have shown that even with a decline in average emergency response time, the average EMV travel time increases from 7.2 minutes in 2015 to 10.1 minutes in 2021 in New York City. Needless to say, there is a severe urgency and significant benefit for shortening the average EMV travel time on increasingly crowded roads.

Existing works have studied strategies to reduce the travel time of EMVs by route optimization and traffic signal pre-emption. Route optimization usually refers to the search for a time-based shortest path. The traffic network (e.g., city road map) is modeled as a graph with intersections as nodes and road segments between intersections as edges. Based on the time a vehicle needs to travel through each edge (road segment), route optimization calculates an optimal route such that an EMV can travel from the rescue station to the incident site in the least amount of time. In addition, as the EMV needs to be as fast as possible, the law in most places requires non-EMVs to yield to emergency vehicles sounding sirens, regardless of the traffic signals at intersections. Even though this practice gives the right-of-way to EMVs, it poses safety risks for vehicles and pedestrians at the intersections. To address this safety concern, existing methods have also studied traffic signal pre-emption which refers to the process of deliberately altering the signal phases at each intersection to prioritize EMV passage.
%
%

However, as the traffic condition constantly changes, an optimal route returned by route optimization can potentially become suboptimal as an EMV travels through the network. Moreover, traffic signal pre-emption has a significant impact on the traffic flow, which would change the fastest route as well. Thus, the optimal route should be updated with real-time traffic flow information, i.e., the route optimization should be solved in a dynamic (time-dependent) way. As an optimal route can change as an EMV travels through the traffic network, the traffic signal pre-emption would need to adapt accordingly. In other words, the subproblems of route optimization and traffic signal pre-emption are coupled and should be solved simultaneously in real-time. Existing approaches does not address this coupling.

%
In addition, most of the existing models on emergency vehicle service have a single objective of reducing the EMV travel time. As a result, their traffic signal control strategies have an undesirable effect of increasing the travel time of non-EMVs, since only EMV passage is optimized. In this paper, we aim to perform route optimization and traffic signal pre-emption to not only reduce EMV travel time but also to reduce the average travel time of non-EMVs. In particular, we address the following two key challenges:
\begin{itemize}[noitemsep]
    \item \textbf{How to dynamically route an EMV to a destination under time-dependent traffic conditions in a computationally efficient way?} As the congestion level of each road segment changes over time, the routing algorithm should be able to update the remaining route as the EMV passes each intersection. Running the shortest-path algorithm each time the EMV passes through an intersection is not efficient. A computationally efficient dynamic routing algorithm is desired. 
    \item \textbf{How to coordinate traffic signals to not only reduce EMV travel time but reduce the average travel time of non-EMVs as well?} To reduce EMV travel time, only the traffic signals along the route of the EMV need to be altered. However, to further reduce average non-EMV travel time, traffic signals in the whole traffic network need to be operated cooperatively.
\end{itemize}

To tackle these challenges, we propose \textbf{EMVLight}, a decentralized multi-agent reinforcement learning framework with a dynamic routing algorithm to control traffic signal phases for efficient EMV passage. Our experimental results demonstrate that EMVLight outperforms traditional traffic engineering methods and existing RL methods under two metrics - EMV travel time and the average travel time of all vehicles - on different traffic configurations.

\section{Related Work}\label{sec_related_work}

\textbf{Conventional routing optimization and traffic signal pre-emption for EMVs.}
%
%
Although, in reality, routing and pre-emption are coupled, the existing methods usually solve them separately. Many of the existing approaches leverage Dijkstra's shortest path algorithm to get the optimal route \cite{wang2013development, Mu2018Route, kwon2003route, JOTSHI20091}. An A* algorithm for ambulance routing has been proposed by \citet{nordin2012finding}. However, as this line of work assumes that the routes and traffic conditions are fixed and static, they fail to address the dynamic nature of real-world traffic flows. Another line of work has considered the change of traffic flows over time. \citet{ziliaskopoulos1993time} have proposed a shortest-path algorithm for time-dependent traffic networks, but the travel time associated with each edge at each time step is assumed to be known in prior. \citet{musolino2013travel} propose different routing strategies for different times in a day (e.g., peak/non-peak hours) based on traffic history data at those times. However, in the problem of our consideration, routing and pre-emption strategies can significantly affect the travel time associated with each edge during the EMV passage, and the existing methods cannot deal with this kind of real-time changes. \citet{haghani2003optimization} formulated the dynamic shortest path problem as a mixed-integer programming problem. \citet{koh2020real} have used RL for real-time vehicle navigation and routing. However, both of these studies have tackled a general routing problem, and signal pre-emption and its influence on traffic have not been modeled.

Once an optimal route for the EMV has been determined, traffic signal pre-emption is deployed. A common pre-emption strategy is to extend the green phases of green lights to let the EMV pass each intersection along a fixed optimal route \cite{wang2013development, bieker2019modelling}. \citet{Asaduzzaman2017APriority} have proposed pre-emption strategies for multiple EMV requests.
%
%

Please refer to \citet{Lu2019Literature} and \citet{humagain2020systematic} for a thorough survey of conventional routing optimization and traffic signal pre-emption methods. We would also like to point out that the conventional methods prioritize EMV passage and have significant disturbances on the traffic flow which increases the average non-EMV travel time.
%
%
%
%

\textbf{RL-based traffic signal control.} 
Traffic signal pre-emption only alters the traffic phases at the intersections where an EMV travels through. However, to reduce congestion, traffic phases at nearby intersections also need to be changed cooperatively. The coordination of traffic signals to mitigate traffic congestion is referred to as traffic signal control which has been addressed by leveraging deep RL in a growing body of work. Many of the existing approaches use Q-learning \cite{abdulhai2003reinforcement, prashanth2010reinforcement, wei2019presslight, wei2019colight, zheng2019frap, ThousandLights}. \citet{Zang_Yao_Zheng_Xu_Xu_Li_2020} leverage meta-learning algorithms to speed up Q-learning for traffic signal control. Another line of work has used actor-critic algorithms for traffic signal control \cite{el2013multiagent, aslani2017adaptive, chu2019multi}. \citet{xu2021hierarchically} propose a hierarchical actor-critic method to encourage cooperation between intersections. Please refer to \citet{wei2019survey} for a review on traffic signal control methods. However, these RL-based traffic control methods focus on reducing the congestion in the traffic network and are not designed for EMV pre-emption. In contrast, our RL framework is built upon state-of-the-art ideas such as max pressure and is designed to reduce both EMV travel time and overall congestion. 
\section{Preliminaries}
  \begin{figure}[b]
    \includegraphics[width=\linewidth]{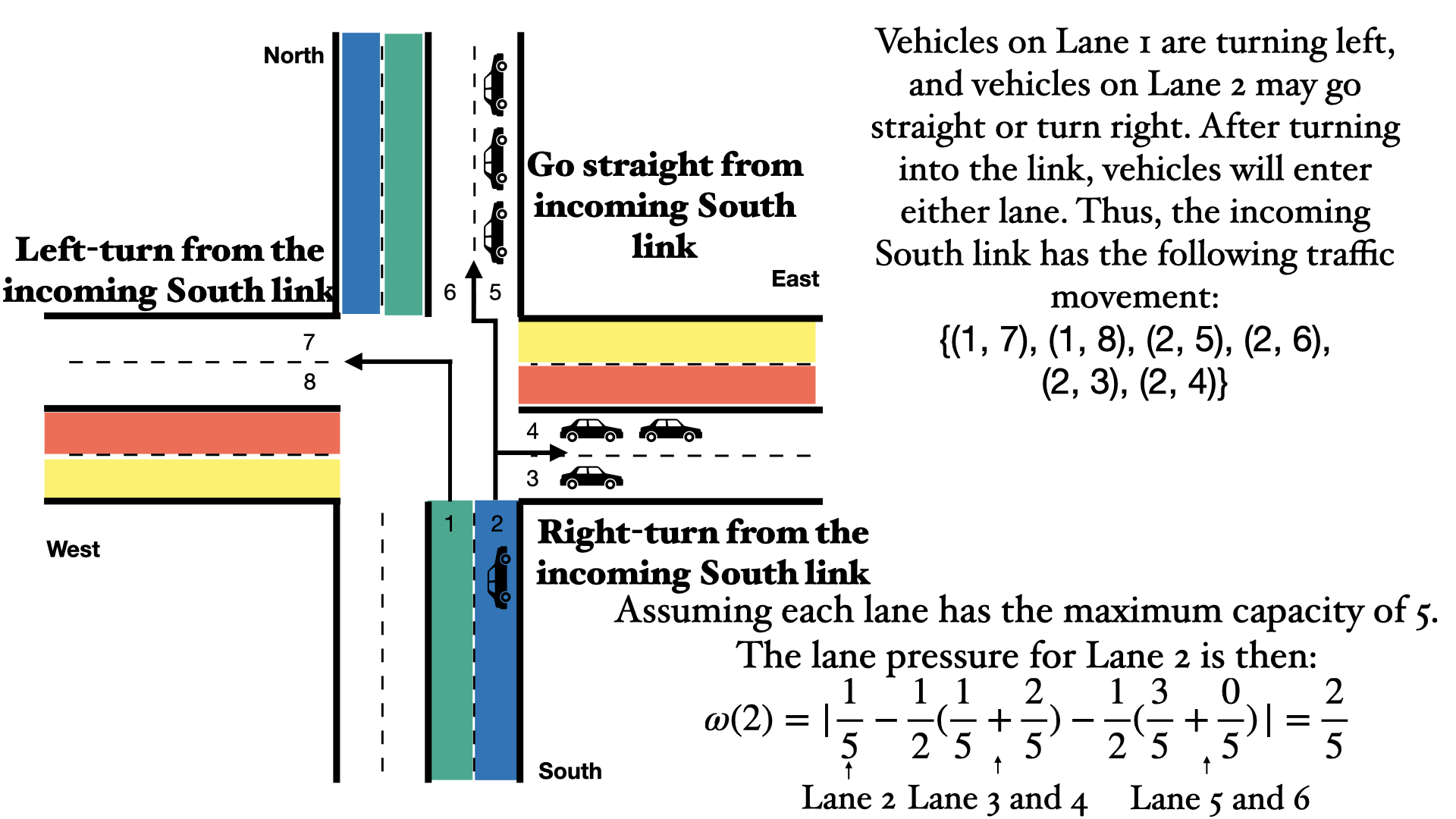}
    \centering
  \caption{Traffic movements illustration and an example pressure calculation for incoming lane \#2.}
  \label{fig_movements}
  \end{figure}
\begin{figure}[t]
    \includegraphics[width=\linewidth]{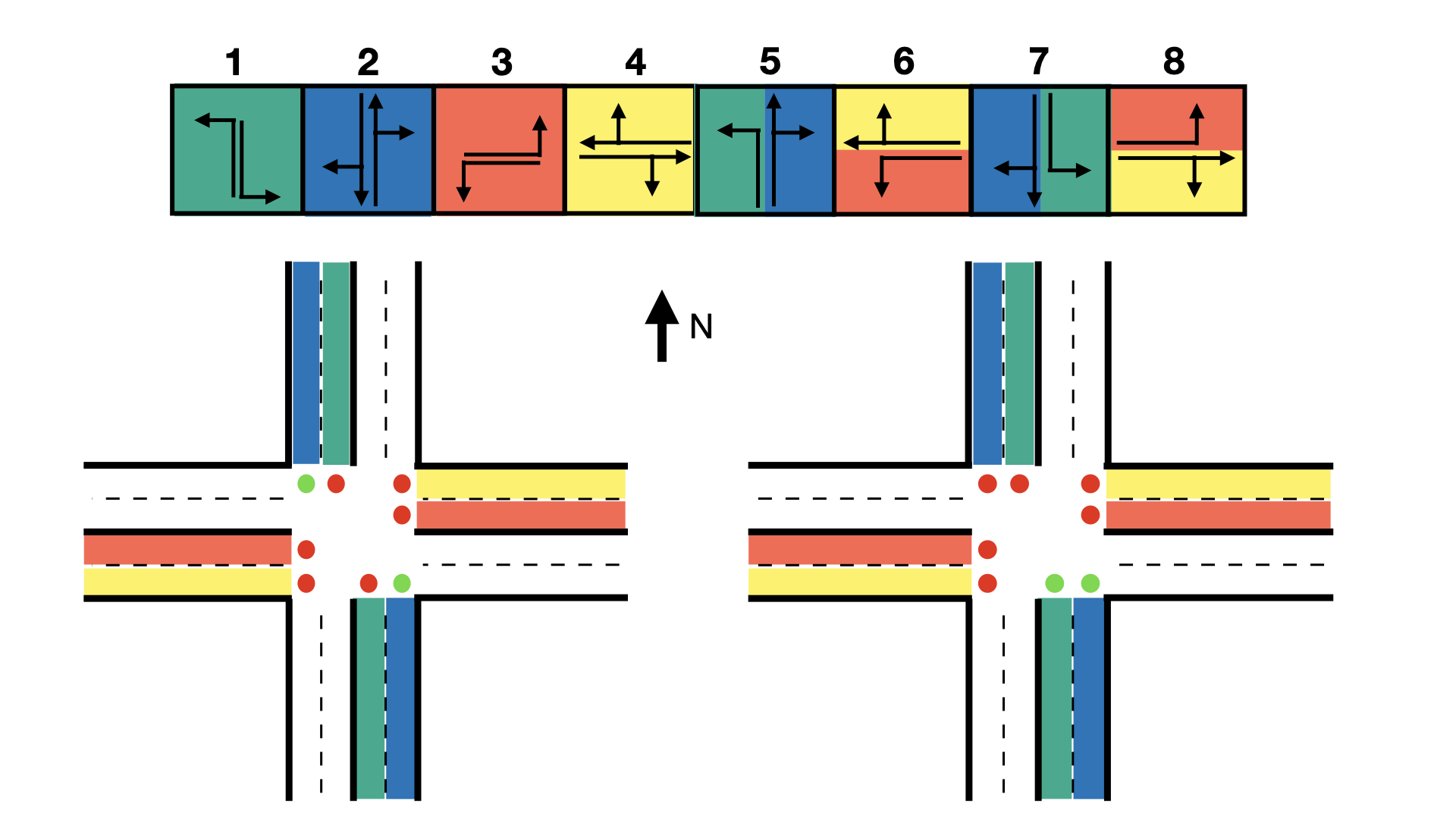}
    \centering
  \caption{\emph{Top}: 8 signal phases; \emph{Left}: phase \#2 illustration; \emph{Right}: phase \#5 illustration.
  }
  \label{fig_phases}
  \end{figure}

\begin{definition}[traffic map, link, lane]
    A traffic map can be represented by a graph $G(\mathcal{V}, \mathcal{E})$, with intersections as nodes and road segments between intersections as edges. We refer to a one-directional road segment between two intersections as a link. A link has a fixed number of lanes, denoted as $h(l)$ for lane $l$. Fig.~\ref{fig_movements} shows 8 links and each link has 2 lanes. 
\end{definition}
\begin{definition}[Traffic movements]
    A traffic movement $(l,m)$ is defined as the traffic traveling across an intersection from an incoming lane $l$ to an outgoing lane $m$. The intersection shown in Fig.~\ref{fig_movements} has 24 permissible traffic movements. The set of all permissible traffic movements of an intersection is denoted as $\mathcal{M}$.
\end{definition}
\begin{definition}[Traffic signal phase]
A traffic signal phase is defined as the set of permissible traffic movements. 
As shown in Fig.~\ref{fig_phases}, an intersection with 4 links has 8 phases.
\end{definition}
%
%
\begin{definition}[Pressure of an incoming lane]
    The pressure of an incoming lane $l$ measures the unevenness of vehicle density between lane $l$ and corresponding out going lanes in permissible traffic movements. The vehicle density of a lane is $x(l)/x_{max}(l)$, where $x(l)$ is the number of vehicles on lane $l$ and $x_{max}(l)$ is the vehicle capacity on lane $l$, which is related to the length of a lane. Then the pressure of an incoming lane $l$ is 
    \begin{equation} 
        w(l) = \left|\frac{x(l)}{x_{max}(l)} - \sum_{\{m|(l, m)\in \mathcal{M}\}}\frac{1}{h(m)}\frac{x(m)}{x_{max}(m)}\right|,
        \label{eqn:lane_pressure}
    \end{equation}
    where $h(m)$ is the number of lanes of the outgoing link which contains $m$. In Fig.~\ref{fig_movements}, $h(m)=2$ for all the outgoing lanes. An example for Eqn.~\eqref{eqn:lane_pressure} is shown in Fig.~\ref{fig_movements}.
\end{definition}

\begin{definition}[Pressure of an intersection]\label{def_intersection_pressure}
    The pressure $P$ of an intersection is the average of the pressure of all incoming lanes.
\end{definition}
The pressure of an intersection indicates the unevenness of vehicle density between incoming and outgoing lanes in an intersection. Intuitively, reducing the pressure leads to more evenly distributed traffic, which indirectly reduce congestion and average travel time of vehicles.

\section{Dynamic Routing}
Dijkstra's algorithm is an algorithm that finds shortest path between a given node and every other nodes in a graph, which has been used for EMV routing. The EMV travel time along each link is estimated based on the number of vehicles on that link. We refer to it as the \emph{intra-link travel time.} Dijkstra's algorithm takes as input the traffic graph, the intra-link travel time and a destination, and can return the time-based shortest path as well as estimated travel time from each intersection to the destination. The latter is usually referred to as the \emph{estimated time of arrival} (ETA) of each intersection.



However, traffic conditions are constantly changing and so does EMV travel time along each link. Moreover, EMV pre-emption techniques alters traffic signal phases, which will significantly change the traffic condition as the EMV travels. The pre-determined shortest path might become congested due to stochasticity and pre-emption. Thus, updating the optimal route dynamically can facilitate EMV passage. In theory we can run Dijkstra's algorithm frequently as the EMV travels through the network to take into account the updated EMV intra-link travel time, but this is inefficient.

To achieve dynamics routing, we extend Dijkstra's algorithm to efficiently update the optimal route based on the updated intra-link travel times. As shown in Algorithm~\ref{alg:ETA_prepopulation}, first a prepopulation process is carried out where a (static) Dijkstra's algorithm is run to get the ETA from each intersection to the destination. For each intersection, the next intersection along the shortest path is also calculated and stored. We assume this process can be done before the EMV starts to travel. This is reasonable since a sequence of processes, including call-taker processing, are performed before the EMVs are dispatched. Once the pre-population process is finished, we can update $\mathsf{ETA}$ and $\mathsf{Next}$ for each intersection efficiently in parallel, since the update only depends on information of neighboring intersections. Please see Appendix for how intra-link travel time is estimated in real time. 

%
\begin{algorithm}[t]
    \caption{Dynamic Dijkstra's for EMV routing}
    \label{alg:ETA_prepopulation}
    \SetEndCharOfAlgoLine{}
    \SetKwInOut{Input}{Input}
    \SetKwInOut{Output}{Output}
    \SetKwData{ETA}{ETA}
    \SetKwData{Next}{Next}
    \SetKwFor{ParrallelForEach}{foreach}{do (in parallel)}{endfor}
    \Input{\\\hspace{-3.7em}
        \begin{tabular}[t]{l @{\hspace{3.3em}} l}
        $G=(\mathcal{V}, \mathcal{E})$ & traffic map as a graph \\
        $T^t = [T_{ij}^t]$ & intra-link travel time at time $t$ \\
        $i_d$  & index of the destination
        \end{tabular}
    }
    \Output{\\\hspace{-3.7em}
        \begin{tabular}[t]{l @{\hspace{1.5em}} l}
        $\mathsf{ETA}^t = [\mathsf{ETA}^t_i]$ & ETA of each intersection \\
        $\mathsf{Next}^t = [\mathsf{Next}^t_i]$ & next intersection to go \\
        & from each intersection
        \end{tabular}
    }
    \tcc{pre-population}
    $\mathsf{ETA}^0, \mathsf{Next}^0$ $=$ \texttt{Dijkstra}$(G, T^0, i_d)$\;
    \tcc{dynamic routing}
    \For{$t = 0 \to T$}{
        \ParrallelForEach{$i \in \mathcal{V}$}{
            $\mathsf{ETA}_i^{t+1} \gets \min_{(i, j)\in \mathcal{E}} (\mathsf{ETA}_j^t + T_{ji}^t)$\;
            $\mathsf{Next}_i^{t+1} \gets \arg\min_{\{j|(i, j)\in \mathcal{E}\}}(\mathsf{ETA}_j^{t} + T_{ji}^t$)\;}}
\end{algorithm}
\begin{remark}
    In static Dijkstra's algorithm, the shortest path is obtained by repeatedly query the $\mathsf{Next}$ attribute of each node from the origin until we reach the destination. In our dynamic Dijkstra's algorithm, since the shortest path changes, at a intersection $i$, we only care about the immediate next intersection to go to, which is exactly $\mathsf{Next}_i$.
\end{remark}
\section{Reinforcement Learning Formulation}
While dynamic routing directs the EMV to the destination, it does not take into account the possible waiting times for red lights at the intersections. Thus, traffic signal pre-emption is also required for the EMV to arrive at the destination in the least amount of time. However, since traditional pre-emption only focuses on reducing the EMV travel time, the average travel time of non-EMVs can increase significantly. Thus, we set up traffic signal control for efficient EMV passage as a decentralized RL problem. In our problem, an RL agent controls the traffic signal phases of an intersection based on local information. Multiple agents coordinate the control signal phases of intersections cooperatively to \textbf{(1)} reduce EMV travel time and \textbf{(2)} reduce the average travel time of non-EMVs. First we design 3 agent types. Then we present agent design and multi-agent interactions.

\subsection{Types of agents for EMV passage}
\begin{figure}[t]
    \centering
    \includegraphics[width=\linewidth]{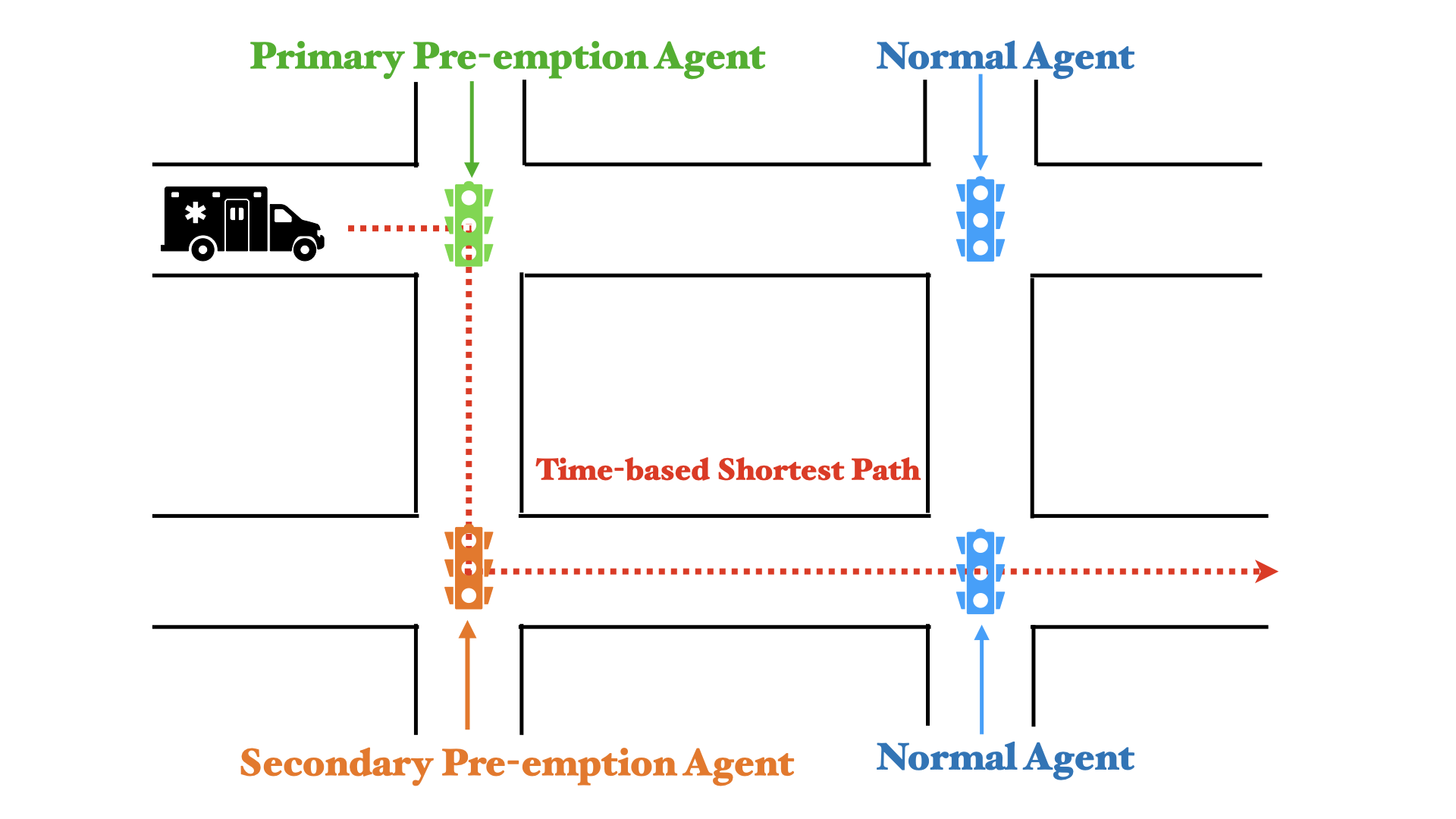}
  \caption{Three types of agents.}
  \label{fig_secondary}
\end{figure}

When an EMV is on duty, we distinguish 3 types of traffic control agents based on EMV location and routing (Fig.~\ref{fig_secondary}). An agent is a \emph{primary pre-emption agent} $i_p$ if an EMV is on one of its incoming links. The agent of the next intersection $i_s = \mathsf{Next}_{i_p}$ is refered to as a \emph{secondary pre-emption agent}.
The rest of the agents are \emph{normal agents}. We design these types since different agents have different local goals, which is reflected in their reward designs. 

\subsection{Agent design}
\begin{itemize}
    \item \textbf{State}: The state of an agent $i$ at time $t$ is denoted as $s^t_i$ and it includes the number of vehicles on each outgoing lanes and incoming lanes, the distance of the EMV to the intersection, the estimated time of arrival ($\mathsf{ETA}$), and which link the EMV will be routed to ($\mathsf{Next}$), i.e.,
    \begin{equation}
        s^t_i = \{x^t(l), x^t(m),  d^t_{\text{EMV}}[L_{ji}], \mathsf{ETA}^{t'}_i, \mathsf{Next}^{t'}_i \},
    \end{equation}
    where $L_{ji}$ represents the links incoming to intersection  $i$, and with a slight abuse of notation $l$ and $m$ denote the set of incoming and outgoing lanes, respectively. For the intersection shown in Fig.~\ref{fig_movements}, $d^t_{\text{EMV}}$ is a vector of four elements. For primary pre-emption agents, one of the elements represents the distance of EMV to the intersection in the corresponding link. The rest of the elements are set to -1. For all other agents, $d^t_{\text{EMV}}$ are padded with -1. 
    \item \textbf{Action}: Prior work has focused on using phase switch, phase duration and phase itself as actions. In this work, we define the action of an agent as one of the 8 phases in Fig.~\ref{fig_phases}; this enables more flexible signal patterns as compared to the traditional cyclical patterns. 
    Due to safety concerns, once a phase has been initiated, it should remain unchanged for a minimum amount of time, e.g. 5 seconds. Because of this, we set our MDP time step length to be 5 seconds to avoid rapid switch of phases. 
    \item \textbf{Reward}: PressLight has shown that minimizing the pressure is an effective way to encourage efficient vehicle passage, we adopt similar idea for normal agents. For secondary pre-emption agents we additionally encourage less vehicle on the link where the EMV is about to enter in order to encourage efficient EMV passage. For primary pre-emption agents, we simply assign a unit penalty at each time step to encourage fast EMV passage. Thus, depending on the agent type, the local reward for agent $i$ at time $t$ is
\end{itemize}
\begin{equation}
    \label{eqn:reward}
    r_{i}^{t} = \begin{cases}
      -P_{i_p}^{t} & i \notin \{i_p, i_s\},\\
      - \beta P_{i_s}^{t} - \frac{1-\beta}{|L_{i_pi_s}|}\sum\limits_{l\in L_{i_pi_s}} \frac{x(l)}{x_{max}(l)}  & i=i_s,\\
      -1 & i=i_p. 
    \end{cases} 
\end{equation}
\textbf{Justification of agent design.} The quantities in local agent state can be obtained at each intersection using various technologies. Numbers of vehicles on each lane $(x^t(l), x^t(m))$ can be obtained by vehicle detection technologies, such as inductive loop \cite{gajda2001vehicle} based on the hardware installed underground. The distance of the EMV to the intersection $d^t_{EMV}[L_{ji}]$ can be obtained by \emph{vehicle-to-infrastructure} technologies such as VANET\cite{buchenscheit2009vanet}, which broadcasts the real-time position of a vehicle to an intersection. Prior work by \citet{wang2013design} and  \citet{noori2016connected} have explored these technologies for traffic signal pre-emption. 

The dynamic routing algorithm (Algorithm~\ref{alg:ETA_prepopulation}) can provide $(\mathsf{ETA}, \mathsf{Next})$ for each agent at every time step. However, due to the stochastic nature of traffic flows, updating the route too frequently might confuse the EMV driver, since the driver might be instructed a new route, say, every 5 seconds. 
There are many ways to ensure reasonable frequency. One option is to inform the driver only once while the EMV is travels in a single link. We implement it by updating the state of an RL agent $(\mathsf{ETA}^{t'}_i, \mathsf{Next}^{t'}_i)$ at the time step when the EMV travels through half of a link. For example, if the EMV travels through a link to agent $i$ from time step 11 to 20 in constant speed, then dynamic routing information in $s_i^{16}$ to $s_i^{20}$ are the same, which is $(\mathsf{ETA}_i^{15}, \mathsf{Next}_i^{15})$, i.e., $t'=15$.

As for the reward design, one might wonder how an agent can know its type. As we assume an agent can observe the state of its neighbors, agent type can be inferred from the observation. This will become clearer below.

\subsection{Multi-agent Advantage Actor-critic}
We adopt a multi-agent advantage actor-critic (MA2C) framework similar to \citet{chu2019multi}. The difference is that our local state includes dynamic routing information and our local reward encourages efficient passage of EMV. Here we briefly introduce the MA2C framework. Please refer to \citet{chu2019multi} for additional details.

In a multi-agent network $G(\mathcal{V}, \mathcal{E})$, the neighborhood of agent $i$ is denoted as $\mathcal{N}_i = \{ j | ji\in \mathcal{E} \textrm{ or } ij\in \mathcal{E}\}$. The local region of agent $i$ is $\mathcal{V}_i = \mathcal{N}_i \cup i$. We define the distance between two agents $d(i, j)$ as the minimum number of edges that connect them. For example, $d(i, i) = 0$ and $d(i, j)=1, \forall j \in \mathcal{N}_i$. In MA2C, each agent learns a policy $\pi_{\theta_i}$ (actor) and the corresponding value function $V_{\phi_i}$ (critic), where ${\theta_i}$ and ${\phi_i}$ are learnable neural network parameters of agent $i$.

\textbf{Local Observation.} In an ideal setting, agents can observe the states of every other agent and leverage this global information to make a decision. However, this is not practical in our problem due to communication latency and will cause scalability issues. We assume agents can observe its own state and the states of its neighbors, i.e., $s^t_{\mathcal{V}_i} = \{s^t_j|j\in \mathcal{V}_i\}$. The agents feed this observation to its policy network $\pi_{\theta_i}$ and value network $V_{\phi_i}$.


\textbf{Fingerprint.} In multi-agent training, each agent treats other agents as part of the environment, but the policy of other agents are changing over time. \citet{foerster2017stabilising} introduce \emph{fingerprints} to inform agents about the changing policies of neighboring agents in multi-agent Q-learning. \citet{chu2019multi} bring fingerprints into MA2C. Here we use the probability simplex of neighboring policies $\pi^{t-1}_{\mathcal{N}_i} = \{\pi^{t-1}_j|j\in \mathcal{N}_i\}$ as fingerprints, and include it into the input of policy network and value network. Thus, our policy network can be written as $\pi_{\theta_i}(a_i^t|s^t_{\mathcal{V}_i}, \pi^{t-1}_{\mathcal{N}_i})$ and value network as $V_{\phi_i}(\Tilde{s}^t_{\mathcal{V}_i}, \pi^{t-1}_{\mathcal{N}_i})$, where $\Tilde{s}^t_{\mathcal{V}_i}$ is the local observation with spatial discount factor, which is introduced below. 

\textbf{Spatial Discount Factor and Adjusted Reward.} MA2C agents cooperatively optimize a global cumulative reward. We assume the global reward is decomposable as $r_t = \sum_{i\in \mathcal{V}} r^t_i$, where $r^t_i$ is defined in Eqn.~\eqref{eqn:reward}. Instead of optimizing the same global reward for every agent, \citet{chu2019multi} propose a spatial discount factor $\alpha$ to let each agent pay less attention to rewards of agents far away. The adjusted reward for agent $i$ is 
\begin{equation}
    \Tilde{r}_i^t = \sum_{d=0}^{D_i}\Big( \sum_{j\in\mathcal{V}|d(i, j)=d} (\alpha)^d r^t_j\Big),
\end{equation}
where $D_i$ is the maximum distance of agents in the graph from agent $i$. When $\alpha > 0$, the adjusted reward include global information, it seems this is in contradiction to the local communication assumption. However, since reward is only used for offline training, global reward information is allowed. Once trained, the RL agents can control traffic signal without relying on global information. 

\textbf{Temporal Discount Factor and Return.} 
The local return $\Tilde{R}^t_i$ is defined as the cumulative adjusted reward $\Tilde{R}^t_i := \sum_{\tau=t}^T \gamma^{\tau-t} \Tilde{r}^\tau_i$, where $\gamma$ is the temporal discount factor and $T$ is the length of an episode. we can estimate the local return using value function,
\begin{equation}
    \Tilde{R}^t_i = \Tilde{r}^t_i + \gamma V_{\phi_i^-}(\Tilde{s}^{t+1}_{\mathcal{V}_i}, \pi^{t}_{\mathcal{N}_i}|\pi_{\theta_{-i}^-}),
\end{equation}
where $\phi_i^-$ means parameters $\phi_i$ are frozen and $\theta_{-i}^-$ means the parameters of policy networks of all other agents are frozen. 

\textbf{Network architecture and training.} As traffic flow data are spatial temporal, we leverage a long-short term memory (LSTM) layer along with fully connected (FC) layers for policy network (actor) and value network (critic). Our multi-agent actor-critic training pipeline is similar to that in \citet{chu2019multi}. We provide neural architecture details, policy loss expression,  value loss expression as well as a training pseudocode in the Appendix.

\section{Experimentation}
In this section, we demonstrate our RL framework using Simulation of Urban MObility (SUMO) \cite{lopez2018microscopic}
SUMO is an open-source traffic simulator capable of simulating both microscopic and macroscopic traffic dynamics, suitable for capturing the EMV's impact on the regional traffic as well as monitoring the overall traffic flow. A pipeline is established between the proposed RL framework and SUMO, i.e., the agents collects observations from SUMO and  preferred signal phases are fed back into SUMO.

\subsection{Datasets and Maps Descriptions}
\begin{figure}[h]
    \centering
    \includegraphics[width=0.9\linewidth]{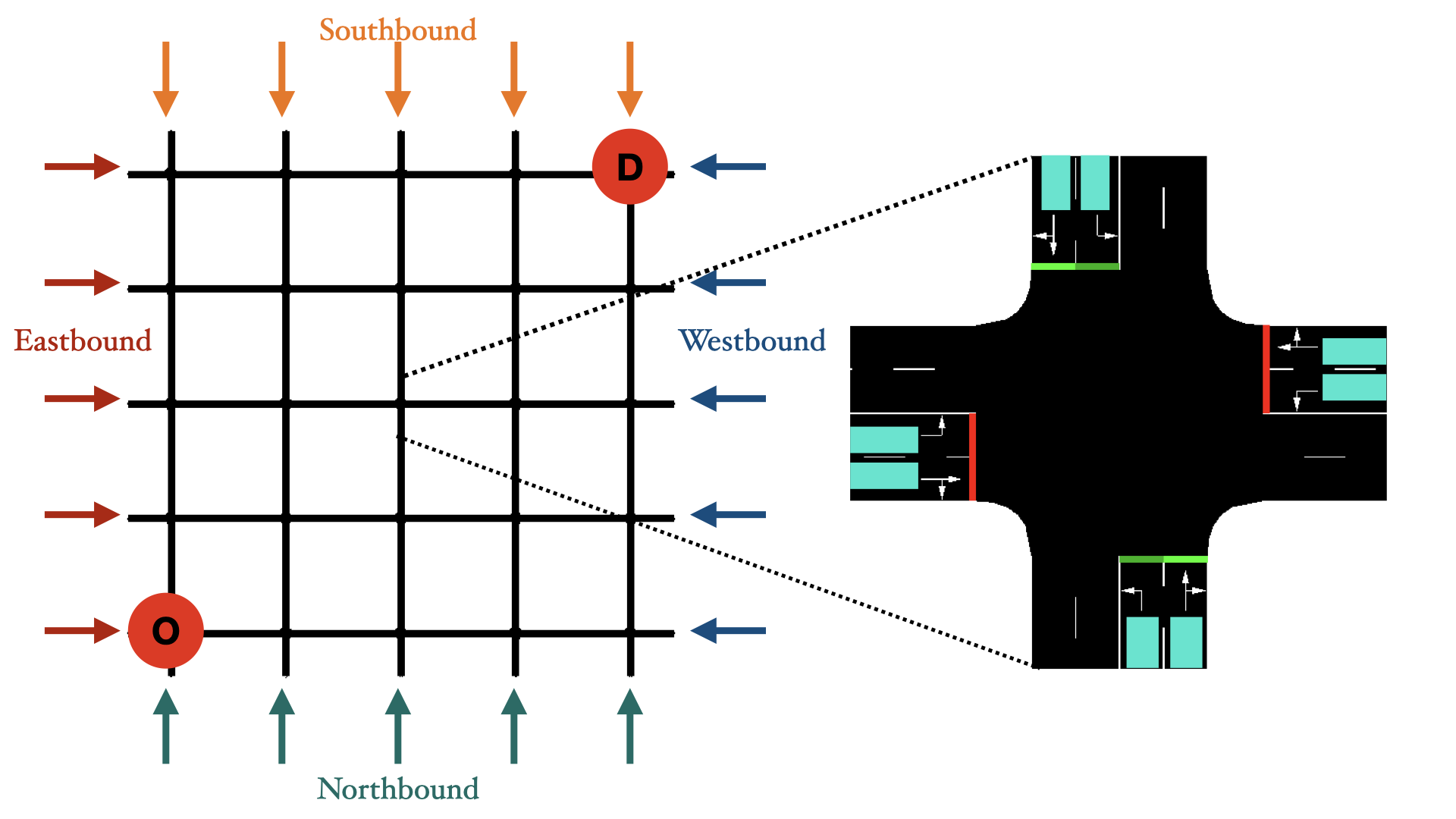}
  \caption{\emph{Left}: the synthetic $\text{grid}_{5\times 5}$. \emph{Right}: an intersection illustration in SUMO, the teal area are inductive loop detected area. Origin and destination for EMV are labeled.}
  \label{fig_synthetic_map}
\end{figure}

We conduct the following experiments based on both synthetic and real-world map. 
\paragraph{Synthetic $\text{Grid}_{5\times 5}$} We synthesize a $5 \times 5$ traffic grid, where intersections are connected with bi-directional links. Each link contains two lanes. We design 4 configurations, listed in Table \ref{tab_synthetic_configuration}. 
The origin (O) and destination (D) of the EMV are labelled in Fig.~\ref{fig_synthetic_map}.
The traffic for this map has a time span of 1200s. We dispatch the EMV at $t =600s$ to ensure the roads are compacted when it starts travel.
\begin{table}[h]
\centering
\fontsize{9.0pt}{10.0pt} \selectfont
\begin{tabular}{@{}ccccc@{}}
\toprule[1pt]
\multicolumn{1}{c}{\multirow{2}{*}{Config}} & \multicolumn{2}{l}{Traffic Flow (veh/lane/hr)} & \multirow{2}{*}{Origin}                                                            & \multirow{2}{*}{Destination}                                                     \\ \cmidrule(lr){2-3}
\multicolumn{1}{c}{}                               & Non-peak                 & Peak                &                                                                                    &                                                                                  \\ 
\cmidrule{1-5}
1                                                  & 200                      & 240                 & \multirow{2}{*}{N,S} & \multirow{2}{*}{E,W} \\
\cmidrule{1-3}
2                                                  & 160                      & 320                 &                                                                                    &                                                                                  \\
\cmidrule{1-5}
3                                                  & 200                      & 240                 & \multicolumn{2}{c}{Randomly}                                                                                                               \\
\cmidrule{1-3}
4                                                  & 160                      & 320                 & \multicolumn{2}{c}{generated}     \\ \bottomrule[1pt]                                                                                                                                   
\end{tabular}
\caption{Configuration for Synthetic $\text{Grid}_{5\times 5}$. Peak flow is assigned from 400s to 800s and non-peak flow is assigned out of this period. For Config. 1 and 2, the vehicles enter the grid from North and South, and exit toward East and West.}
\label{tab_synthetic_configuration}
\end{table}

\begin{figure}[h]
    \centering
    \includegraphics[width=\linewidth]{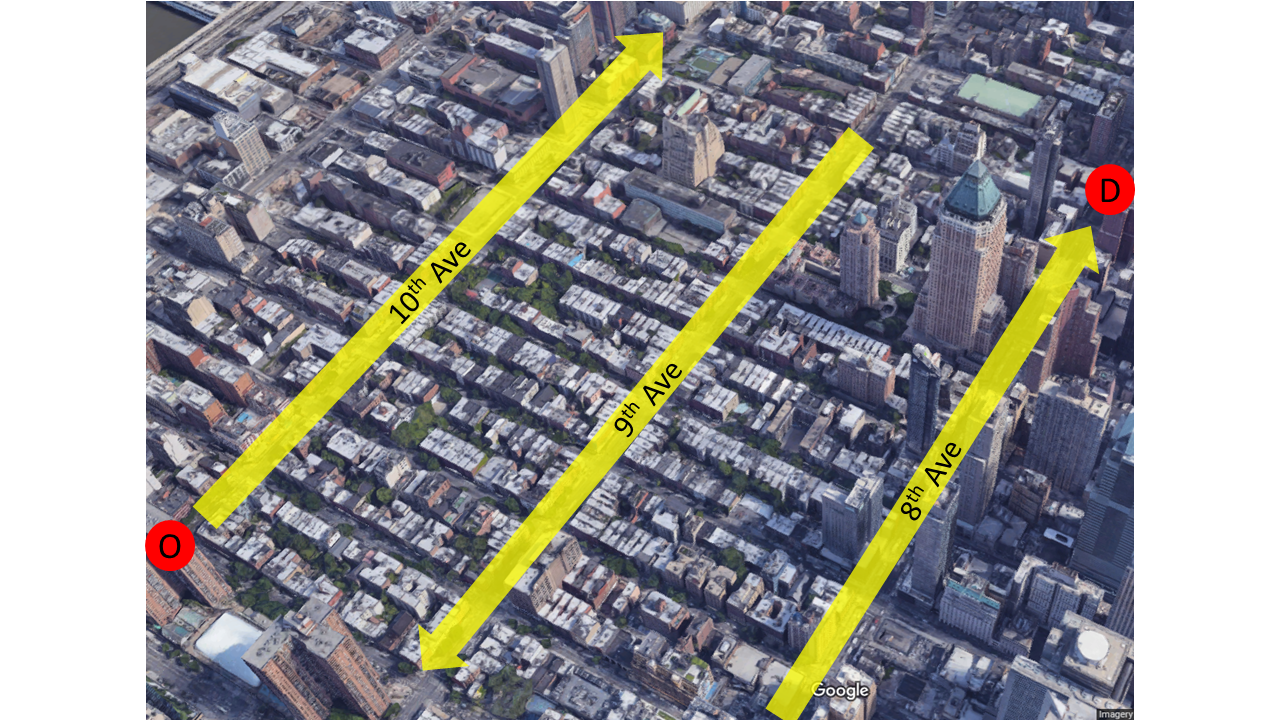}
  \caption{Manhattan map: a 16-by-3 traffic network in Hell's Kitchen area. Origin and destination for the EMV dispatching are labeled.}
  \label{fig_manhattan}
\end{figure}

\paragraph{$\text{Manhattan}_{16\times 3}$}
This is a $16 \times 3$ traffic network extracted from Manhattan Hell's Kitchen area (Fig.~\ref{fig_manhattan}) and customized for demonstrating EMV passage. In this traffic network, intersections are connected
by 16 one-directional streets and 3 one-directional avenues. We assume each avenue contains four lanes and each street contains two lanes so that the right-of-way of EMVs and pre-emption can be demonstrated. The traffic flow for this map is generated from open-source NYC taxi data. Both the map and traffic flow data are publicly available.\footnote{https://traffic-signal-control.github.io/}
The origin and destination of EMV are set to be far away as shown in Fig.~\ref{fig_manhattan}
\begin{table*}[t]
\centering
\fontsize{9.0pt}{10.0pt} \selectfont
\setlength\tabcolsep{3.4pt}
\begin{tabular}{@{}cccccc|ccccc@{}}
\toprule[1pt]
\multirow{2}{*}{Method
} & \multicolumn{5}{c|}{EMV Travel Time $T_{\textrm{EMV}}$ [s]}                                  & \multicolumn{5}{c}{Average Travel Time $T_{\textrm{avg}}$ [s]}                                             \\ \cmidrule(l){2-11} 
                        & Config 1     & Config 2     & Config 3     & Config 4     & $\text{Manhattan}_{16\times 3}$    & Config 1        & Config 2        & Config 3        & Config 4        & $\text{Manhattan}_{16\times 3}$       \\ \midrule
FT w/o EMV          & N/A          & N/A          & N/A          & N/A          & N/A          & 353.43          & 371.13          & 314.25          & 334.10          & 1649.64       \\ \midrule
W + Static + FT             & 257.20         & 272.00          & 259.20          & 243.80          & 487.20          & 372.19          & 389.13          & 342.49          & 355.05          & 1811.03         \\
W + Static + MP             & 255.00          & 269.00          & 261.20          & 245.40          & 461.80          & 349.38          & 352.54          & 307.91          & 322.68          & 708.13          \\
W + Static + CL           & 281.20          & 286.20          & 289.80          & 277.80          & 492.20          & 503.35          & 524.26          & 488.12          & 509.55          & 2013.54         \\
W + Static + PL             & 276.00          & 282.20          & 271.40          & 275.00          & 476.00          & 358.18          & 369.45          & 332.98          & 338.95          & 1410.76         \\ \midrule
W + dynamic + FT                 & 229.60          & 231.20          & 228.60          & 227.20          & 442.20          & 370.09          & 393.40          & 330.13          & 345.50          & 1699.30         \\
W + dynamic + MP                 & 226.20          & 234.60          & 224.20          & 217.60          & 438.80          & 345.45          & 348.43          & 313.26          & 325.72          & 721.32          \\
W + dynamic + CL                & 273.40          & 269.60          & 281.00          & 270.80          & 450.20          & 514.29          & 536.78          & 502.12          & 542.63          & 1987.86         \\
W + dynamic + PL                 & 251.20         & 257.80          & 247.00          & 268.80          & 436.20          & 359.31          & 342.59          & 340.11          & 349.20          & 1412.12         \\ \midrule
EMVLight                & \textbf{198.60} & \textbf{192.20} & \textbf{199.20} & \textbf{196.80} & \textbf{391.80} & \textbf{322.40} & \textbf{318.76} & \textbf{301.90} & \textbf{321.02} & \textbf{681.23} \\ \bottomrule[1pt]
\end{tabular}%
\caption{Performance comparison of different methods evaluated in the four configurations of the synthetic traffic grid as well as Manhattan Map. For both metrics, the lower value indicates better performance. The lowest values are highlighted in bold. The average travel time of Manhattan map (1649.64) is retrieved from data. 
}
\label{tab_performance_review}
\end{table*}
\subsection{Baselines}
Due to the lack of existing RL methods for efficient EMV passage, we select traditional methods and RL methods for each subproblem and combine them to set up baselines. 

For traffic signal pre-emption, the most intuitive and widely used approach is extending green light period for EMV passage at each intersection which results in a \emph{Green Wave} \cite{corman2009evaluation}. 
\textbf{Walabi (W)} \cite{bieker2019modelling} is 
an effective rule-based method that implemented Green Wave for EMVs in SUMO environment. We integrate Walabi with combinations of routing and traffic signal control strategies introduced below as baselines.

\emph{Routing baselines:}
\begin{itemize}
    \item \textbf{Static} routing is performed when EMV is dispatched and the route remains fixed as the EMV travels. We adopt A* search as the baseline since it is an powerful extension to the Dijkstra's shortest path algorithm and is used in many real-time applications because of its optimality. \footnote{Our implementation of A* search employs a Manhattan distance as the heuristic function.}
    \item \textbf{Dynamic} routing relies on real-time information of traffic conditions. To set up the baseline, we run A* every 50s as EMV travels. This is because running the full A* to update optimal route is not as efficient as our proposed dynamic Dijkstra's algorithm. 

\end{itemize}

\emph{Traffic signal control baselines:}
\begin{itemize}
    \item \textbf{Fixed Time (FT)}: Cyclical fixed time traffic phases with random offset \cite{roess2004traffic} is a policy that split all phases with an predefined green ratio. 
    It is the default strategy in real traffic signal control.
    \item \textbf{Max Pressure (MP)}: The state-of-the-art (SOTA) network-level signal control strategy based on pressure \cite{varaiya2013max}. It aggressively select the phase with maximum pressure to smooth congestion.
    \item \textbf{Coordinated Learner (CL)}: A Q-learning based coordinator which directly learns joint local value functions for adjacent intersections \cite{van2016coordinated}.
    \item \textbf{PressLight (PL)}: A RL method aiming to optimize the pressure at each intersection\cite{wei2019presslight}.
\end{itemize}

\subsection{Results}
We evaluate performance of models under two metrics: \emph{EMV travel time}, which reflects routing and pre-emption ability, and \emph{average travel time}, which indicates the ability of traffic signal control for efficient vehicle passage.  The performance of our EMVLight and the baselines in both the synthetic and the Manhattan map is shown in Table \ref{tab_performance_review}. 
The results of all methods are averaged over five independent runs and RL methods are tested with random seeds. We observe that EMVLight outperforms all baseline models under both metrics. 

In terms of EMV travel time $T_{\textrm{EMV}}$, the dynamic routing baseline performs better than static routing baselines. This is expected since dynamic routing considers the time-dependent nature of traffic conditions and update optimal route accordingly. EMVLight further reduces EMV travel time by 18\% in average as compared to dynamic routing baselines. This advantage in performance can be attributed to the design of secondary pre-emption agents. This type of agents learn to ``reserve a link" by choosing signal phases that help clear the vehicles in the link to encourage high speed EMV passage (Eqn.~\eqref{eqn:reward}).

As for average travel time $T_{\textrm{avg}}$, we first notice that the traditional pre-emption technique (W+Static+FT) indeed increases the average travel time by around 10\% as compared to a traditional Fix Time strategy without EMV (denoted as ``FT w/o EMV" in Table \ref{tab_performance_review}), thus decreasing the efficiency of vehicle passage. Different traffic signal control strategies have a direct impact on overall efficiency. Fixed Time is designed to handle steady traffic flow. Max Pressure, as a SOTA traditional method, outperforms Fix Time and, surprisingly, outperforms both RL baselines in terms of overall efficiency. This shows that pressure is an effective indicator for reducing congestion and this is why we incorporate pressure in our reward design. Coordinate Learner performs the worst probably because its reward is not based on pressure. PressLight doesn't beat Max Pressure because it has a reward design that focuses on smoothing vehicle densities along a major direction, e.g. an arterial. Grid networks with the presence of EMV make PressLight less effective. Our EMVLight improves its pressure-based reward design to encourage smoothing vehicle densities of all directions for each intersection. This enable us to achieve an advantage of 5\% over our best baselines (Max Pressure).

\subsubsection{Ablation study on pressure and agent types}
We propose three types of agents and design their rewards (Eqn.~\eqref{eqn:reward}) based on our improved pressure definition and heuristics. 
In order to see how our improved pressure definition and proposed special agents influence the results, we (1) replace our pressure definition by that defined in PressLight, (2) replace secondary pre-emption agents with normal agents and (3) replace primary pre-emption agents with normal agents. 
\begin{table}[b]
\centering
\fontsize{9.0pt}{10.0pt} \selectfont
\begin{tabular}{@{}cccc|c@{}}
\toprule[1pt]
Ablations                        & (1) & (2) & (3) & EMVLight\\ \midrule
$T_{\text{EMV}}$ [s]         & \textbf{197}       & 289                     & 320     & 199       \\
$T_{\text{avg}}$ [s]  & 361.05    & 347.13                  & 359.62   & \textbf{322.40}      \\ \bottomrule[1pt]
\end{tabular}
\caption{Ablation study on pressure and agent types. Experiments are conducted on the Config 1 synthetic $\text{grid}_{5 \times 5}$.}
\label{tab_ablation_reward}
\end{table}

Table \ref{tab_ablation_reward} shows the results of these ablations: (1) PressLight-style pressure (see Appendix) yields a slightly smaller EMV travel time but significantly increases the average travel time; (2) Without secondary pre-emption agents, EMV travel time increases by 45\% since almost no ``link reservation" happened; (3) Without primary pre-emption agents, EMV travel time increases significantly, which shows the importance of pre-emption.

\subsubsection{Ablation study on fingerprint}
\begin{figure}[ht]
    \centering
    \includegraphics[width=0.9\linewidth]{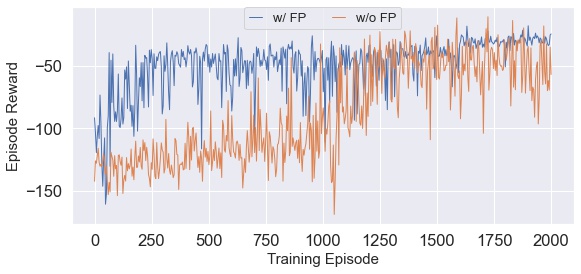}
  \caption{Reward convergence with and without fingerprint. Experiments are conducted on Config 1 synthetic $\text{grid}_{5 \times 5}$.}
  \label{fig_FP_comparison}
\end{figure}

In multi-agent RL, fingerprint has been shown to stabilize training and enable faster convergence.
In order to see how fingerprint affects training in EMVLight, we remove the fingerprint design, i.e., policy and value networks are changed from $\pi_{\theta_i}(a_i^t|s^t_{\mathcal{V}_i}, \pi^{t-1}_{\mathcal{N}_i})$ and  $V_{\phi_i}(\Tilde{s}^t_{\mathcal{V}_i}, \pi^{t-1}_{\mathcal{N}_i})$ to $\pi_{\theta_i}(a_i^t|s^t_{\mathcal{V}_i})$ and  $V_{\phi_i}(\Tilde{s}^t_{\mathcal{V}_i})$, respectively. 
Fig.~\ref{fig_FP_comparison} shows the influence of fingerprint on training. With fingerprint, the reward converges faster and suffers from less fluctuation, confirming the effectiveness of fingerprint. 
\section{Conclusion}
In this paper, we proposed a decentralized reinforcement learning framework, EMVLight, to facilitate the efficient passage of EMVs and reduce traffic congestion at the same time.
Leveraging the multi-agent A2C framework, agents incorporate dynamic routing and cooperatively control traffic signals to reduce EMV travel time and average travel time of non-EMVs. 
Evaluated on both synthetic and real-world map, EMVLight significantly outperforms the existing methods.
Future work will explore more realistic microscopic interaction between EMV and non-EMVs, efficient passage of multiple EMVs and closing the sim-to-real gap.


\bibliography{main}

\begin{thebibliography}{37}
\providecommand{\natexlab}[1]{#1}

\bibitem[{Abdulhai, Pringle, and Karakoulas(2003)}]{abdulhai2003reinforcement}
Abdulhai, B.; Pringle, R.; and Karakoulas, G.~J. 2003.
\newblock Reinforcement learning for true adaptive traffic signal control.
\newblock \emph{Journal of Transportation Engineering}, 129(3): 278--285.

\bibitem[{Analytics(2021)}]{end-to-end-response-times}
Analytics, N. 2021.
\newblock End-To-End Response Times.
\newblock
  \url{https://www1.nyc.gov/site/fdny/about/resources/data-and-analytics/end-to-end-response-times.page}.

\bibitem[{{Asaduzzaman} and {Vidyasankar}(2017)}]{Asaduzzaman2017APriority}
{Asaduzzaman}, M.; and {Vidyasankar}, K. 2017.
\newblock A Priority Algorithm to Control the Traffic Signal for Emergency
  Vehicles.
\newblock In \emph{2017 IEEE 86th Vehicular Technology Conference (VTC-Fall)},
  1--7.

\bibitem[{Aslani, Mesgari, and Wiering(2017)}]{aslani2017adaptive}
Aslani, M.; Mesgari, M.~S.; and Wiering, M. 2017.
\newblock Adaptive traffic signal control with actor-critic methods in a
  real-world traffic network with different traffic disruption events.
\newblock \emph{Transportation Research Part C: Emerging Technologies}, 85:
  732--752.

\bibitem[{Bieker-Walz and Behrisch(2019)}]{bieker2019modelling}
Bieker-Walz, L.; and Behrisch, M. 2019.
\newblock Modelling green waves for emergency vehicles using connected traffic
  data.
\newblock \emph{EPiC Series in Computing}, 62: 1--11.

\bibitem[{Buchenscheit et~al.(2009)Buchenscheit, Schaub, Kargl, and
  Weber}]{buchenscheit2009vanet}
Buchenscheit, A.; Schaub, F.; Kargl, F.; and Weber, M. 2009.
\newblock A VANET-based emergency vehicle warning system.
\newblock In \emph{2009 IEEE Vehicular Networking Conference (VNC)}, 1--8.
  IEEE.

\bibitem[{Chen et~al.(2020)Chen, Wei, Xu, Zheng, Yang, Xiong, Xu, and
  Li}]{ThousandLights}
Chen, C.; Wei, H.; Xu, N.; Zheng, G.; Yang, M.; Xiong, Y.; Xu, K.; and Li, Z.
  2020.
\newblock Toward A Thousand Lights: Decentralized Deep Reinforcement Learning
  for Large-Scale Traffic Signal Control.
\newblock \emph{Proceedings of the AAAI Conference on Artificial Intelligence},
  34(04): 3414--3421.

\bibitem[{Chu et~al.(2019)Chu, Wang, Codec{\`a}, and Li}]{chu2019multi}
Chu, T.; Wang, J.; Codec{\`a}, L.; and Li, Z. 2019.
\newblock Multi-Agent Deep Reinforcement Learning for Large-Scale Traffic
  Signal Control.
\newblock \emph{IEEE Transactions on Intelligent Transportation Systems}.

\bibitem[{Corman et~al.(2009)Corman, D’Ariano, Pacciarelli, and
  Pranzo}]{corman2009evaluation}
Corman, F.; D’Ariano, A.; Pacciarelli, D.; and Pranzo, M. 2009.
\newblock Evaluation of green wave policy in real-time railway traffic
  management.
\newblock \emph{Transportation Research Part C: Emerging Technologies}, 17(6):
  607--616.

\bibitem[{El-Tantawy, Abdulhai, and Abdelgawad(2013)}]{el2013multiagent}
El-Tantawy, S.; Abdulhai, B.; and Abdelgawad, H. 2013.
\newblock Multiagent reinforcement learning for integrated network of adaptive
  traffic signal controllers (MARLIN-ATSC): methodology and large-scale
  application on downtown Toronto.
\newblock \emph{IEEE Transactions on Intelligent Transportation Systems},
  14(3): 1140--1150.

\bibitem[{Foerster et~al.(2017)Foerster, Nardelli, Farquhar, Afouras, Torr,
  Kohli, and Whiteson}]{foerster2017stabilising}
Foerster, J.; Nardelli, N.; Farquhar, G.; Afouras, T.; Torr, P.~H.; Kohli, P.;
  and Whiteson, S. 2017.
\newblock Stabilising experience replay for deep multi-agent reinforcement
  learning.
\newblock In \emph{International conference on machine learning}, 1146--1155.
  PMLR.

\bibitem[{Gajda et~al.(2001)Gajda, Sroka, Stencel, Wajda, and
  Zeglen}]{gajda2001vehicle}
Gajda, J.; Sroka, R.; Stencel, M.; Wajda, A.; and Zeglen, T. 2001.
\newblock A vehicle classification based on inductive loop detectors.
\newblock In \emph{IMTC 2001. Proceedings of the 18th IEEE Instrumentation and
  Measurement Technology Conference. Rediscovering Measurement in the Age of
  Informatics}, volume~1, 460--464. IEEE.

\bibitem[{Haghani, Hu, and Tian(2003)}]{haghani2003optimization}
Haghani, A.; Hu, H.; and Tian, Q. 2003.
\newblock An optimization model for real-time emergency vehicle dispatching and
  routing.
\newblock In \emph{82nd annual meeting of the Transportation Research Board,
  Washington, DC}. Citeseer.

\bibitem[{Humagain et~al.(2020)Humagain, Sinha, Lai, and
  Ranjitkar}]{humagain2020systematic}
Humagain, S.; Sinha, R.; Lai, E.; and Ranjitkar, P. 2020.
\newblock A systematic review of route optimisation and pre-emption methods for
  emergency vehicles.
\newblock \emph{Transport reviews}, 40(1): 35--53.

\bibitem[{Jotshi, Gong, and Batta(2009)}]{JOTSHI20091}
Jotshi, A.; Gong, Q.; and Batta, R. 2009.
\newblock Dispatching and routing of emergency vehicles in disaster mitigation
  using data fusion.
\newblock \emph{Socio-Economic Planning Sciences}, 43(1): 1 -- 24.

\bibitem[{Koh et~al.(2020)Koh, Zhou, Fang, Yang, Yang, Yang, Guan, and
  Ji}]{koh2020real}
Koh, S.; Zhou, B.; Fang, H.; Yang, P.; Yang, Z.; Yang, Q.; Guan, L.; and Ji, Z.
  2020.
\newblock Real-time deep reinforcement learning based vehicle navigation.
\newblock \emph{Applied Soft Computing}, 96: 106694.

\bibitem[{Kwon, Kim, and Betts(2003)}]{kwon2003route}
Kwon, E.; Kim, S.; and Betts, R. 2003.
\newblock Route-based dynamic preemption of traffic signals for emergency
  vehicle operations.
\newblock In \emph{Transportation Research Board 82nd Annual
  MeetingTransportation Research Board}.

\bibitem[{Lopez et~al.(2018)Lopez, Behrisch, Bieker-Walz, Erdmann,
  Fl{\"o}tter{\"o}d, Hilbrich, L{\"u}cken, Rummel, Wagner, and
  Wie{\ss}ner}]{lopez2018microscopic}
Lopez, P.~A.; Behrisch, M.; Bieker-Walz, L.; Erdmann, J.; Fl{\"o}tter{\"o}d,
  Y.-P.; Hilbrich, R.; L{\"u}cken, L.; Rummel, J.; Wagner, P.; and Wie{\ss}ner,
  E. 2018.
\newblock Microscopic traffic simulation using sumo.
\newblock In \emph{2018 21st International Conference on Intelligent
  Transportation Systems (ITSC)}, 2575--2582. IEEE.

\bibitem[{{Lu} and {Wang}(2019)}]{Lu2019Literature}
{Lu}, L.; and {Wang}, S. 2019.
\newblock Literature Review of Analytical Models on Emergency Vehicle Service:
  Location, Dispatching, Routing and Preemption Control.
\newblock In \emph{2019 IEEE Intelligent Transportation Systems Conference
  (ITSC)}, 3031--3036.

\bibitem[{Mu, Song, and Liu(2018)}]{Mu2018Route}
Mu, H.; Song, Y.; and Liu, L. 2018.
\newblock Route-Based Signal Preemption Control of Emergency Vehicle.
\newblock \emph{Journal of Control Science and Engineering}, 2018: 1--11.

\bibitem[{Musolino et~al.(2013)Musolino, Polimeni, Rindone, and
  Vitetta}]{musolino2013travel}
Musolino, G.; Polimeni, A.; Rindone, C.; and Vitetta, A. 2013.
\newblock Travel time forecasting and dynamic routes design for emergency
  vehicles.
\newblock \emph{Procedia-Social and Behavioral Sciences}, 87: 193--202.

\bibitem[{Noori, Fu, and Shiravi(2016)}]{noori2016connected}
Noori, H.; Fu, L.; and Shiravi, S. 2016.
\newblock A connected vehicle based traffic signal control strategy for
  emergency vehicle preemption.
\newblock In \emph{Transportation Research Board 95th Annual Meeting}, 16-6763.

\bibitem[{Nordin et~al.(2012)Nordin, Zaharudin, Maasar, and
  Nordin}]{nordin2012finding}
Nordin, N. A.~M.; Zaharudin, Z.~A.; Maasar, M.~A.; and Nordin, N.~A. 2012.
\newblock Finding shortest path of the ambulance routing: Interface of A-star
  algorithm using C programming.
\newblock In \emph{2012 IEEE Symposium on Humanities, Science and Engineering
  Research}, 1569--1573. IEEE.

\bibitem[{Prashanth and Bhatnagar(2010)}]{prashanth2010reinforcement}
Prashanth, L.; and Bhatnagar, S. 2010.
\newblock Reinforcement learning with function approximation for traffic signal
  control.
\newblock \emph{IEEE Transactions on Intelligent Transportation Systems},
  12(2): 412--421.

\bibitem[{Roess, Prassas, and McShane(2004)}]{roess2004traffic}
Roess, R.~P.; Prassas, E.~S.; and McShane, W.~R. 2004.
\newblock \emph{Traffic engineering}.
\newblock Pearson/Prentice Hall.

\bibitem[{Su et~al.(2021)Su, Shi, Chow, and Jin}]{su2021dynamic}
Su, H.; Shi, K.; Chow, J. Y.~J.; and Jin, L. 2021.
\newblock Dynamic Queue-Jump Lane for Emergency Vehicles under Partially
  Connected Settings: A Multi-Agent Deep Reinforcement Learning Approach.
\newblock arXiv:2003.01025.

\bibitem[{Van~der Pol and Oliehoek(2016)}]{van2016coordinated}
Van~der Pol, E.; and Oliehoek, F.~A. 2016.
\newblock Coordinated deep reinforcement learners for traffic light control.
\newblock \emph{Proceedings of Learning, Inference and Control of Multi-Agent
  Systems (at NIPS 2016)}.

\bibitem[{Varaiya(2013)}]{varaiya2013max}
Varaiya, P. 2013.
\newblock Max pressure control of a network of signalized intersections.
\newblock \emph{Transportation Research Part C: Emerging Technologies}, 36:
  177--195.

\bibitem[{Wang, Ma, and Yang(2013)}]{wang2013development}
Wang, J.; Ma, W.; and Yang, X. 2013.
\newblock Development of degree-of-priority based control strategy for
  emergency vehicle preemption operation.
\newblock \emph{Discrete dynamics in nature and society}, 2013.

\bibitem[{Wang et~al.(2013)Wang, Wu, Yang, and Huang}]{wang2013design}
Wang, Y.; Wu, Z.; Yang, X.; and Huang, L. 2013.
\newblock Design and implementation of an emergency vehicle signal preemption
  system based on cooperative vehicle-infrastructure technology.
\newblock \emph{Advances in Mechanical Engineering}, 5: 834976.

\bibitem[{Wei et~al.(2019{\natexlab{a}})Wei, Chen, Zheng, Wu, Gayah, Xu, and
  Li}]{wei2019presslight}
Wei, H.; Chen, C.; Zheng, G.; Wu, K.; Gayah, V.; Xu, K.; and Li, Z.
  2019{\natexlab{a}}.
\newblock Presslight: Learning max pressure control to coordinate traffic
  signals in arterial network.
\newblock In \emph{Proceedings of the 25th ACM SIGKDD International Conference
  on Knowledge Discovery \& Data Mining}, 1290--1298.

\bibitem[{Wei et~al.(2019{\natexlab{b}})Wei, Xu, Zhang, Zheng, Zang, Chen,
  Zhang, Zhu, Xu, and Li}]{wei2019colight}
Wei, H.; Xu, N.; Zhang, H.; Zheng, G.; Zang, X.; Chen, C.; Zhang, W.; Zhu, Y.;
  Xu, K.; and Li, Z. 2019{\natexlab{b}}.
\newblock Colight: Learning network-level cooperation for traffic signal
  control.
\newblock In \emph{Proceedings of the 28th ACM International Conference on
  Information and Knowledge Management}, 1913--1922.

\bibitem[{Wei et~al.(2019{\natexlab{c}})Wei, Zheng, Gayah, and
  Li}]{wei2019survey}
Wei, H.; Zheng, G.; Gayah, V.; and Li, Z. 2019{\natexlab{c}}.
\newblock A Survey on Traffic Signal Control Methods.
\newblock \emph{arXiv preprint arXiv:1904.08117}.

\bibitem[{Xu et~al.(2021)Xu, Wang, Wang, Jia, and Lu}]{xu2021hierarchically}
Xu, B.; Wang, Y.; Wang, Z.; Jia, H.; and Lu, Z. 2021.
\newblock Hierarchically and Cooperatively Learning Traffic Signal Control.
\newblock In \emph{Proceedings of the AAAI Conference on Artificial
  Intelligence}, volume~35, 669--677.

\bibitem[{Zang et~al.(2020)Zang, Yao, Zheng, Xu, Xu, and
  Li}]{Zang_Yao_Zheng_Xu_Xu_Li_2020}
Zang, X.; Yao, H.; Zheng, G.; Xu, N.; Xu, K.; and Li, Z. 2020.
\newblock MetaLight: Value-Based Meta-Reinforcement Learning for Traffic Signal
  Control.
\newblock \emph{Proceedings of the AAAI Conference on Artificial Intelligence},
  34(01): 1153--1160.

\bibitem[{Zheng et~al.(2019)Zheng, Xiong, Zang, Feng, Wei, Zhang, Li, Xu, and
  Li}]{zheng2019frap}
Zheng, G.; Xiong, Y.; Zang, X.; Feng, J.; Wei, H.; Zhang, H.; Li, Y.; Xu, K.;
  and Li, Z. 2019.
\newblock Learning phase competition for traffic signal control.
\newblock In \emph{Proceedings of the 28th ACM International Conference on
  Information and Knowledge Management}, 1963--1972.

\bibitem[{Ziliaskopoulos and Mahmassani(1993)}]{ziliaskopoulos1993time}
Ziliaskopoulos, A.~K.; and Mahmassani, H.~S. 1993.
\newblock Time-dependent, shortest-path algorithm for real-time intelligent
  vehicle highway system applications.
\newblock In \emph{Transportation Research Record 1408}, 94--100.

\end{thebibliography}

\pagebreak
\renewcommand\appendixpagename{Supplementary Materials}
\appendix
\appendixpage
\renewcommand{\thesection}{S\arabic{section}}

\setcounter{equation}{0}
\renewcommand{\theequation}{S.\arabic{equation}}
\setcounter{figure}{0}
\renewcommand{\thefigure}{S.\arabic{figure}}
\setcounter{table}{0}
\renewcommand{\thetable}{S.\arabic{table}}
\setcounter{algocf}{0}
\renewcommand{\thealgocf}{S.\arabic{algocf}}

\section{Pressure Definition Comparison} 
Here we present the key difference in pressure definition between our work and PressLight \cite{wei2019presslight}.

\subsection{Pressure in PressLight}
PressLight assumes that traffic movements are lane-to-lane, i.e., vehicles in one lane can only move into a particular lane in a link. Because of the lane-to-lane assumption, in PressLight, the pressure is defined per movement. PressLight defines the pressure of a movement as the difference of the vehicle density between an incoming lane $l$ and the outgoing lane $m$, i.e., 
\begin{equation*}
    w^{*}(l, m) = \frac{x(l)}{x_{max}(l)} - \frac{x(m)}{x_{max}(m)},
\end{equation*}

PressLight then defines the pressure of an intersection $i$ as the absolute value of the sum of pressure of movements of intersection $i$, i.e., 
\begin{equation*}\label{eq:PressLight_pressure}
    P^{*}_{i} = |\sum _{(l, m)\in \mathcal{M}_i} w^{*}(l, m)|,
\end{equation*}
where $\mathcal{M}_i$ is the set of permissible traffic movements of intersection $i$.

\subsection{Pressure in our work}
EMVLight assumes a lane-to-link style traffic movement as vehicles can enter either lane on the target link, see Fig 1. To present the pressure of an intersection, we first define the pressure of an incoming lane as (Definition 4)
\begin{equation*} 
    w(l) = \left|\frac{x(l)}{x_{max}(l)} - \sum_{\{m|(l, m)\in \mathcal{M}\}}\frac{1}{h(m)}\frac{x(m)}{x_{max}(m)}\right|.
\end{equation*}
The pressure of an intersection in EMVLight is defined as the average of the pressure of all incoming lanes (Definition 5),
\begin{equation*}\label{eq:EMVLight_pressure}
    P_{i} = \frac{1}{|\mathcal{I}_i|}\sum _{l\in \mathcal{I}_i} w(l),
\end{equation*}
where $\mathcal{I}_i$ represents the set of all incoming lanes of intersection $i$.

\subsection{Comparison}
The first difference between the two definitions is that $w^{*}(l, m)$ can be both positive or negative, but $w(l)$ can only take positive values that measures the unevenness of the vehicle density in the incoming lane and that of the corresponding outgoing lanes. We take the absolute value since the direction of pressure is irrelevant here, and the goal of each agent is to minimize this unevenness. The second difference is that at the intersection level, $P^{*}_{i}$ takes a sum but $P_{i}$ takes an average. The average is more suitable for our purpose since it scales the pressure down and the unit penalty for normal agents would be relatively large as compared to rewards for pre-emption agents (Eqn. (3)). This design puts the efficient passage of EMV vehicles at the top priority. Our experimentation results indicate the proposed pressure design produces a more robust reward signal during training and outperforms PressLight in congestion reduction.

\section{Intralink EMV travel time}

The intra-link traffic pattern with the presence of an EMV on duty is complicated and is under-explored in the current literature. For simplicity, here we demonstrate a simple intra-link traffic model for a link with 2 lanes. The model can be easily extended for multiple lanes.

In a two-lane link, the EMV takes a lane and the non-EMVs on the other lane usually slows down or entirely stop. Some non-EMVs ahead of the EMV find pull-over spots in the other lane and park there. Those that cannot find a parking spot continue to drive in front of the EMV, potentially blocking the EMV passage \cite{su2021dynamic}. In this study, we propose a meso-scopic model to estimate the intra-link travel time of an EMV.
\begin{figure}[h]
    \centering
    \includegraphics[width=\linewidth]{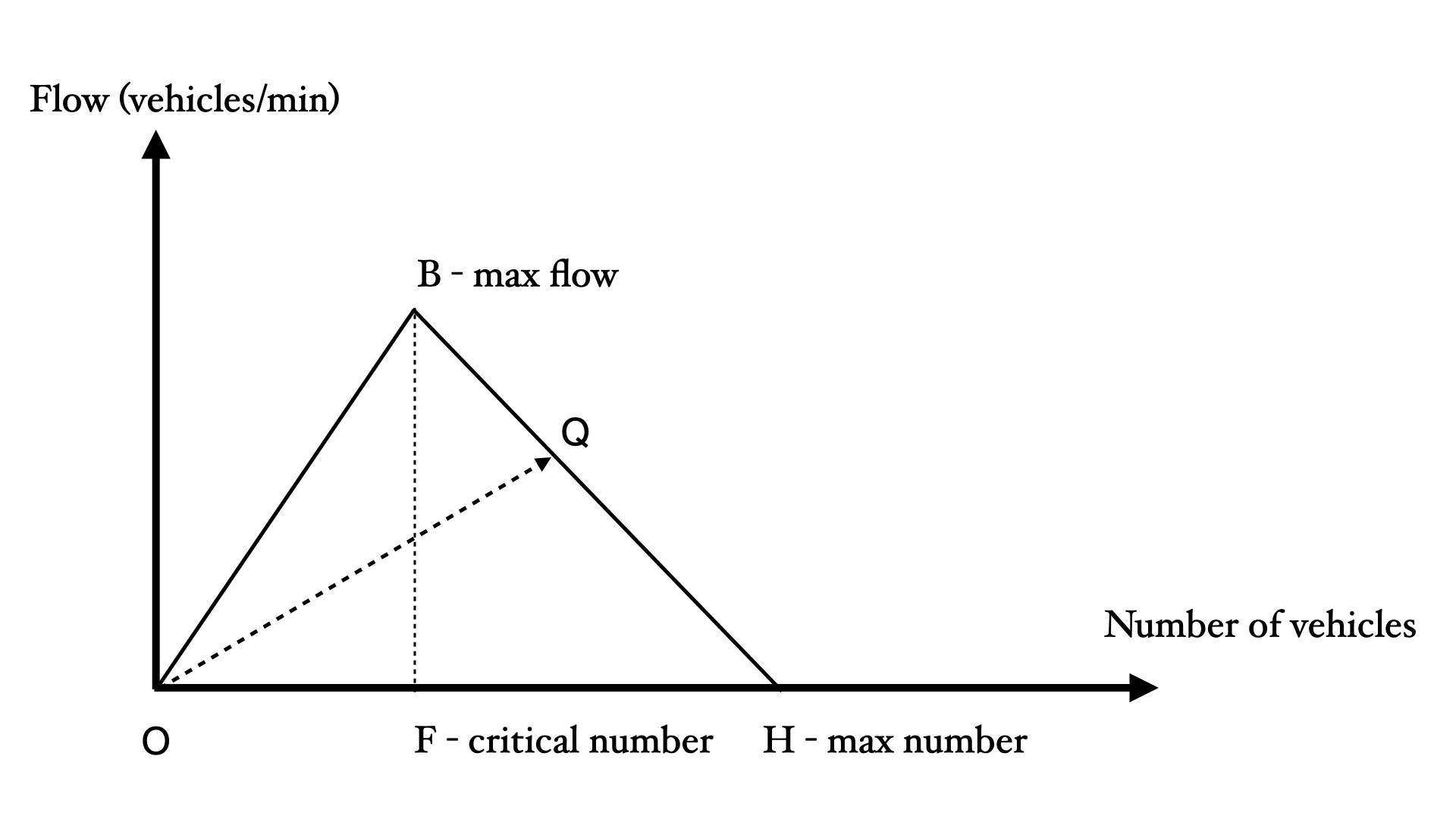}
  \caption{Normal traffic state.}
  \label{fig_two_lanes}
\end{figure}
Normally, the traffic flow of a link is modeled by a fundamental diagram, see Fig \ref{fig_two_lanes}. This diagram depicts a simplified relationship between the flow rate, i.e. number of vehicles passing within the unit amount of the time, and number of vehicles on a link. The max number of vehicles $H$ indicates the capacity of this link. The critical number of vehicles $F$ indicates the boundary differentiates the non-congested state and congested state. When the number of vehicles is smaller than $F$, all vehicles are traveling at the free flow speed, which is represented by the slope of $OB$. When number of vehicles is larger than $F$, vehicles are slowing down and traffic flows declines since the link is now congested. The max flow is attained when the number of vehicles is at $F$. The travel speed of vehicles in a congested state $Q$ is obtained by the slope of $OQ$.
\begin{figure}[h]
    \centering
    \includegraphics[width=\linewidth]{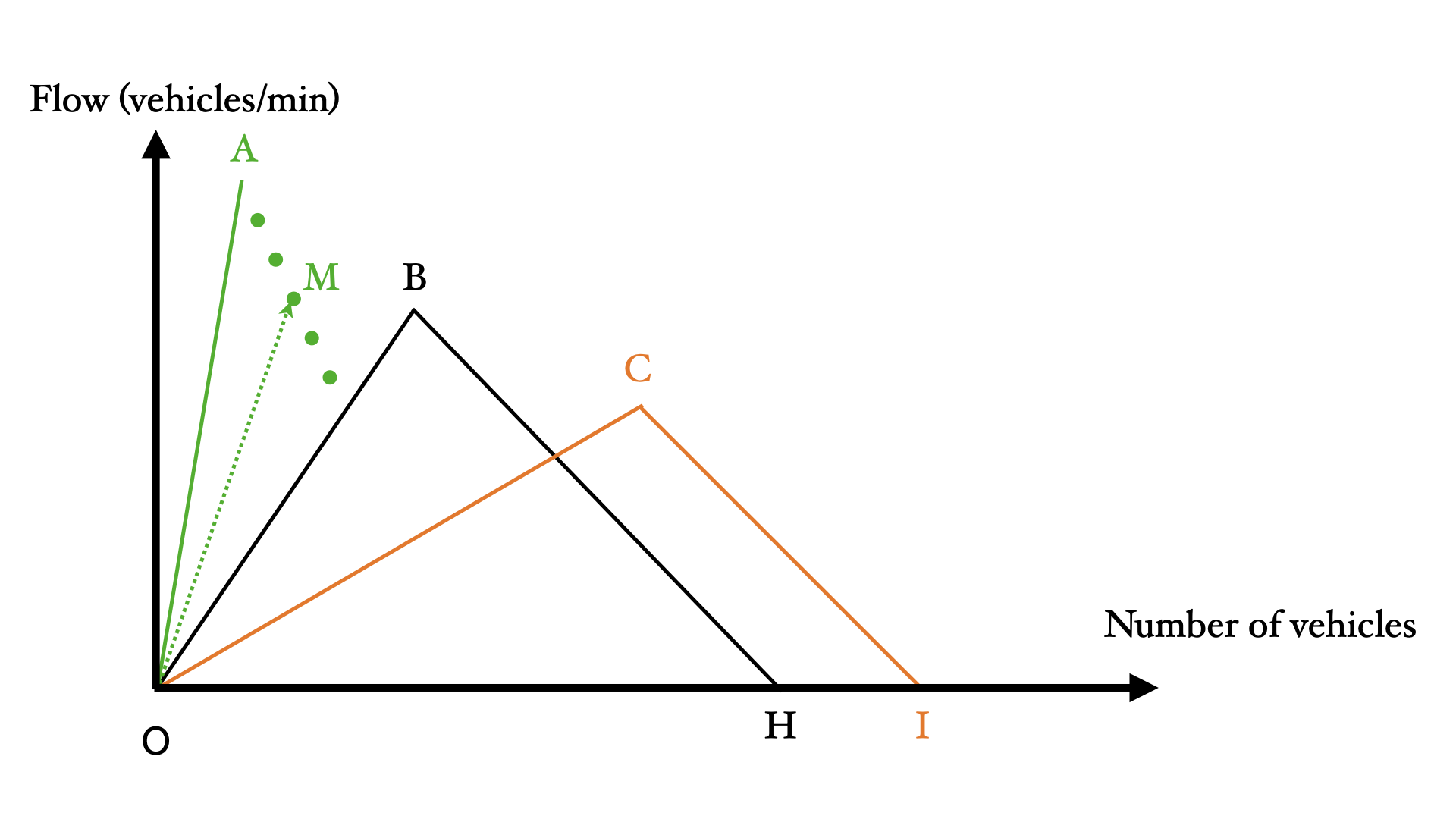}
  \caption{Traffic state during EMV pre-emption.}
  \label{fig_one_lane}
\end{figure}

During the EMV pre-emption, the original traffic flow relationship pictured in black are diverted into two parts, representing two lanes respectively. The green line represents the traffic conditions of the pre-emption lane, i.e. the lane where the EMV is traveling, and the orange line represents the other lane. During pre-emption, part of the vehicles originally travelling in front of the EMV pull over onto the adjacent lane, resulting a significant decrease in the max capacities of the pre-emption lane. Meanwhile, because vehicles can park onto the curbs, orange line depicts a larger maximum capacity $I$ than normal.

Regarding the max flow, the adjacent lane has a smaller max flow as the vehicles on this lane are required to slow down when EMV is on duty.
The pre-emption lane obtains a max capacity when there is one vehicle on the lane, i.e. the EMV itself. Under this circumstance, the flow rate is equivalent as the free flow speed of the EMV. However, when there are vehicles remaining in front of the EMV, the travel speed of the EMV might be slowed down, but still higher than the free flow speeds of the non-EMVs. Furthermore, the travel speed of EMV has a discrete value corresponding to the number of non-EMVs blocking. For example, pre-emption lane has a traffic state represented by $M$, and the travel speed of the EMV is obtained as the slope of $M$. We use this simple model, especially the green plot to estimate the intra-link travel time of EMV in a link as a function of number of vehicles in that link. 

We calibrate this model with the SUMO environment. Intuitively, the travel speed of the EMV is affected more by the number of vehicles on the pre-emption lane than their positions. The reason behind is since the ETA is frequently updated every $\Delta T$ seconds ($\Delta T$ in our experiment), and the estimated ETA would eventually converge. Imagine there are vehicles very far away from the EMV and about to leave the intersections at $t$, their presence would not slow down the approaching EMV. Therefore, when they have left the intersection at $t+1$, we have an updated number of non-EMVs count and updated travel speed estimation.

\begin{algorithm}[h]
    \caption{Multi-agent A2C Training}
    \label{alg:training}
    \SetEndCharOfAlgoLine{}
    \SetKwInOut{Input}{Input}
    \SetKwInOut{Output}{Output}
    \SetKwData{ETA}{ETA}
    \SetKwData{Next}{Next}
    \SetKwFor{ParrallelForEach}{foreach}{do (in parallel)}{endfor}
    \Input{\\\hspace{-3.7em}
        \begin{tabular}[t]{l @{\hspace{3.3em}} l}
        $T$ & maximum time step of an episode \\
        $N_{\mathrm{bs}}$ & batch size \\
        $\eta_\theta$  & learning rate for policy networks \\
        $\eta_\phi$    & learning rate for value networks \\
        $\alpha$       & spatial discount factor \\
        $\gamma$       & (temporal) discount factor\\ $\lambda$      & regularizer coefficient
        \end{tabular}
    }
    \Output{\\\hspace{-3.7em}
        \begin{tabular}[t]{l @{\hspace{1.4em}} l}
        $\{\phi_i\}_{i\in\mathcal{V}}$ & learned parameters in value networks \\
        $\{\theta_i\}_{i\in\mathcal{V}}$ & learned parameters in policy networks \\
        \end{tabular}
    }
    \textbf{initialize} $\{\phi_i\}_{i\in\mathcal{V}}$, $\{\theta_i\}_{i\in\mathcal{V}}$, $k \gets 0$, $B \gets \varnothing$;
    \textbf{initialize} SUMO, $t \gets 0$, \textbf{get} $\{s^0_i\}_{i\in\mathcal{V}}$\;
    \Repeat{Convergence}{
        \tcc{generate trajectories}
        \ParrallelForEach{$i \in \mathcal{V}$}{
            \textbf{sample} $a^t_i$ from $\pi^t_i$\;
            \textbf{receive} $\Tilde{r}^t_i$ and $s^{t+1}_i$\;
        }
        $B \gets B \cup \{(s_i^t, \pi_i^t, a_i^t, s_i^{t+1}, r_i^t)_{i\in \mathcal{V}}\}$\;
        $t \gets t+1$, $k \gets k+1$\;
        \If{$t == T$}{
            \textbf{initialize} SUMO, $t \gets 0$, \textbf{get} $\{s^0_i\}_{i\in\mathcal{V}}$\;
        }
        \tcc{update actors and critics}
        \If{$k == N_{\mathrm{bs}}$}{
            \ParrallelForEach{$i \in \mathcal{V}$}{
                \textbf{calculate} $\Tilde{r}^t_i$ (Eqn. (4)), $\Tilde{R}^t_i$ (Eqn. (5))\;
                $\phi_i \gets \phi_i - \eta_\phi \nabla \mathcal{L}_v(\phi_i)$\;
                $\theta_i \gets \theta_i - \eta_\theta \nabla \mathcal{L}_p(\theta_i)$\;
            }
            $k \gets 0, B \gets \varnothing$\;
        }
    }
    
\end{algorithm}

\section{Training Details}

\subsection{Value loss function}
With a batch of data $B = \{(s_i^t, \pi_i^t, a_i^t, s_i^{t+1}, r_i^t)_{i\in \mathcal{V}}^{t\in \mathcal{T}}\}$, each agent's value network is trained by minimizing the difference between bootstrapped estimated value and neural network approximated value
\begin{equation}
    \label{eqn:L_v}
    \mathcal{L}_v(\phi_i) = \frac{1}{2|B|} \sum_{B}\Big( \Tilde{R}^t_i - V_{\phi_i}(\Tilde{s}^t_{\mathcal{V}_i}, \pi^{t-1}_{\mathcal{N}_i}) \Big)^2.
\end{equation}

\subsection{Policy loss function}
Each agent's policy network is trained by minimizing its policy loss
\begin{align}
    \label{eqn:L_p}
    \mathcal{L}_p(\theta_i) = -& \frac{1}{|B|}\sum_{B} \bigg(\log \pi_{\theta_i}(a_i^t|s^t_{\mathcal{V}_i}, \pi^{t-1}_{\mathcal{N}_i}) \Tilde{A}^t_i \\
    &- \lambda \sum_{a_i \in \mathcal{A}_i} \pi_{\theta_i} \log \pi_{\theta_i} (a_i | s^t_{\mathcal{V}_i}, \pi^{t-1}_{\mathcal{N}_i}) \bigg),
\end{align}
where $\Tilde{A}^t_i = \Tilde{R}^t_i - V_{\phi_i^-}(\Tilde{s}^t_{\mathcal{V}_i}, \pi^{t-1}_{\mathcal{N}_i})$ is the estimated advantage which measures how much better the action $a^t_i$ is as compared to the average performance of the policy $\pi_{\theta_i}$ in the state $s_i^t$. The second term is a regularization term that encourage initial exploration, where $\mathcal{A}_i$ is the action set of agent $i$. For an intersection as shown in Fig. 1, $\mathcal{A}_i$ contains 8 traffic signal phases.

\subsection{Training algorithm}
Algorithm \ref{alg:training} shows the multi-agent A2C training process.

\section{Implementation Details}
\subsection{Implementation details for synthetic $\text{grid}_{5 \times 5}$}
\begin{itemize}
    \item dimension of $s^t_{\mathcal{V}_i}$: $5 \times (8+8+4+2)=110$
    \item dimension of $\Tilde{s}^t_{\mathcal{V}_i}$: $5 \times (8+8+4+2)=110$
    \item dimension of $\pi^{t-1}_{\mathcal{N}_i}$: $4 \times 8 = 32$
    \item Policy network $\pi_{\theta_i}(a_i^t|s^t_{\mathcal{V}_i}, \pi^{t-1}_{\mathcal{N}_i})$: 
    \texttt{concat[}$110 \xrightarrow[]{\textrm{FC}} 128$ReLu, $32 \xrightarrow[]{\textrm{FC}} 64$ReLu\texttt{]} $ \xrightarrow[]{} 64$LSTM $ \xrightarrow[]{\textrm{FC}}8$Softmax
    \item Value network $V_{\phi_i}(\Tilde{s}^t_{\mathcal{V}_i}, \pi^{t-1}_{\mathcal{N}_i})$: \texttt{concat[}$110 \xrightarrow[]{\textrm{FC}} 128$ReLu, $32 \xrightarrow[]{\textrm{FC}} 64$ReLu\texttt{]} $ \xrightarrow[]{} 64$LSTM $ \xrightarrow[]{\textrm{FC}}1$Linear
    \item Each link is $200m$. The free flow speed of the EMV is $12m/s$ and the free flow speed for non-EMVs is $6m/s$.
    \item Temporal discount factor $\gamma$ is $0.99$ and spatial discount factor $\alpha$ is $0.90$.
    \item Initial learning rates $\eta_\phi$ and $\eta_\theta$ are both 1e-3 and they decay linearly. Adam optimizer is used.
    \item MDP step length $\Delta t = 5s$ and for secondary pre-emption reward weight $\beta$ is $0.5$.
    \item Regularization coefficient is $0.01$.
\end{itemize}
\subsection{Implementation details for $\text{Manhattan}_{16 \times 3}$}
The implementation is similar to the synthetic network implementation, with the following differences:
\begin{itemize}
    \item Initial learning rates $\eta_\phi$ and $\eta_\theta$ are both 5e-4.
    \item Since the avenues and streets are both one-directional, the number of actions of each agent are adjusted accordingly. 
\end{itemize}

\end{document}